# Sexism in the Judiciary:
# Bias Definition in NLP and in Our Courts

**Noa Baker Gillis**
Tel Aviv, Israel
noabakergillis@gmail.com

## Abstract

We analyze 6.7 million case law documents to determine the presence of gender bias within our judicial system. We find that current bias detection methods in NLP are insufficient to determine gender bias in our case law database and propose an alternative approach. We show that existing algorithms' inconsistent results are consequences of prior research's inconsistent definitions of biases themselves. Bias detection algorithms rely on groups of words to represent bias (e.g., 'salary,' 'job,' and 'boss' to represent employment as a potentially biased theme against women in text). However, the methods to build these groups of words have several weaknesses, primarily that the word lists are based on the researchers' own intuitions. We suggest two new methods of automating the creation of word lists to represent biases. We find that our methods outperform current NLP bias detection methods. Our research improves the capabilities of NLP technology to detect bias and highlights gender biases present in influential case law. In order to test our NLP bias detection method's performance, we regress our results of bias in case law against U.S census data of women's participation in the workforce in the last 100 years.

## 1 Introduction

Are gender biases present in our judicial system, and can machine learning detect them? Drawing on the idea that text can provide insight into human psychology (Jakiela and Ozier, 2019), we look at gender-stereotyped language in case law as a proxy for bias in our judicial system. Unfortunately, previous NLP work in bias detection is insufficient to robustly determine bias in our database (Zhang et al., 2019). We show that previous bias detection methods all share a common flaw: these algorithms rely on groups of words to represent a potential bias (e.g., 'salary,' 'job,' and 'boss' to represent employment as a potential bias against women) that are not standardized. This lack of standardization is flawed in three main ways. First, these word lists are built by the researchers with little explanation and are susceptible to researchers' own implicit biases. Consequently, the words within the word list might not truly describe the bias as it exists in the text. Second, the same bias theme (e.g., 'employment') often has different word lists in different papers. Inconsistent word lists lead to varied results. As we show, using two different researcher's word lists to represent a bias on a single database can produce almost opposite results. Third, there is little discussion about the method of choosing words to represent specific biases. It is therefore difficult to reproduce or extend existing research on bias detection.

In order to search meaningfully for gender bias within our judicial system, we propose two methods for automatically creating word lists to represent biases in text. We find that our methods outperform existing bias detection methods and we employ our new methods to identify gender bias in case law. We find that this bias exists. Finally, we map gender bias's progress over time and find that bias against women in case law decreases at about the same rate, at the same time, that women enter the workforce in the last 100 years.

## 2 Bias Statement

In this paper, we study gender bias in case law using two new NLP methods. We define gender bias in text as a measurable asymmetry in language when discussing men versus women (excluding group-specific words such as gender pronouns). Bias is especially harmful in the context of case law decisions. If case law systematically associates men more positively and powerfully than women, the law creates representational harm by perpetuating unfair and inaccurate stereotypes. Further, bias in law could lead to failure to account for gender-related harms that could disproportionately affect women. For example, because of the imposition of restrictions on recovery, there is no reliable means of tort compensation for victims of domestic violence, rape, and sexual assault (Chamallas, 2018). This is just one example where failure to equally consider both genders in law leads to real harm.

The proposed bias detection algorithm only detects bias after the case has been written. However, case law is unique in that it sets precedent for other, later cases: judges often cite previous cases as a basis for a new judgment. Therefore, we suggest that this bias detection method be used as a way for judges to more deeply understand biases present in the cases they cite. Perhaps a deeper understanding of biases in historical cases could prevent biases from reappearing in new judgments.

## 3 Related Works

A variety of bias detection methods have been proposed in gender-related literature. Prominent among these methods is the Implicit Associations Test (IAT) (Nosek, Greenwald, and Banaji, 20). IAT measures the strength of associations between groups (e.g., men, women) and evaluations (e.g., good, bad) or stereotypes (e.g., strong, weak) to which those groups are assigned. The main idea is that classifying a group is easier, and therefore happens more quickly, when the subject agrees with the evaluation. For example, a subject has an implicit bias towards men relative to women if they are faster to classify men as strong / women as weak, than women as strong / men as weak.

In NLP literature, the most prominent bias detection method is the Word Embedding Association Test (WEAT). WEAT measures the association of word lists representing a potentially biased theme (e.g., 'salary', 'job,' and 'boss' to represent employment) to a set of pronoun or otherwise gendered pairs such as (she, he) or (man, woman) (Bolukbase et al., 2016; Caliskan et al., 2017; Garg et al., 2018; Freidman et al., 2019). The association is measured by first training a word embedding model on text. The researchers then compute the distance of vectors relating to gendered word pairs (e.g., she / he) to words in word lists representing potential bias categories. The average distance of the words in a themed word list is the magnitude of bias. The vector direction (i.e., the positive or negative distance) represents towards which group the bias is directed. WEAT uses the vector direction and proximity as a proxy for semantic association.

WEAT has been used to uncover biases in many databases. For example, Garg et al (2017) used WEAT to detect bias in Google News. Other research has used WEAT to identify bias in twitter posts in 99 countries (Friedman et al., 2019). One particularly relevant study to our research uses the same database of case law to study gender bias using WEAT, finding that cases written by female and younger judges tend to have less bias than their older, male counterparts (Ash, Chen, and Ornaghi, 2020). Another particularly relevant work uses WEAT to study the presence of gender bias in four different databases (Chaloner and Maldonado, 2019). The same work also suggests a preliminary method of automatically detecting word lists to represent gender bias but falls short of suggesting a way to determine the relevance of each bias category.

The efficacy of WEAT in bias detection is inconsistent. WEAT also fails robustness tests: for example, the average bias magnitude of words in an employment word list might be skewed towards men, but there could be words within the word list whose bias magnitude skews towards women (Zhang et al., 2019). Even different capitalizations of the same word might have different bias magnitudes

## 4 Methodology

### 4.1 Data

We use the Case Law Access Project (CAP) as our dataset. Released by Harvard Law in late 2018, the database contains over 6.7 million unique U.S state case decisions. Case law in the U.S plays a fundamental role in common-law policy making

due to its ability to set precedent, making CAP an influential, rich database for judgment analysis

## 4.2 Overview of Approaches

We propose two new methods of identifying gender bias in case law. The first method employs a word frequency ('first-order') processing algorithm to identify words more frequently used when discussing one gender than the other in our case law database. We group the outputted gendered words thematically and use the resulting word lists, representing biases, as inputs to WEAT. The second approach employs the same first-order processing method to identify bias words. Instead of manually grouping the resulting list of gendered words thematically, we use popular automatic clustering algorithm K-Means. K-Means clustering groups our vectors representing words by proximity and similarity. We use the resulting clusters as inputs to WEAT. We compare the outputs of our methods to existing word group by performing robustness tests described in recent literature and find that both our suggested methods outperform the current standard.

## 4.3 WLOR: First Order Processing

For both approaches, we use a first-order sorting method to identify words used more frequently for women than men in our database. The purpose is to use the resulting most-gendered words for word lists representing biases as inputs to WEAT. We hypothesize that even using this light, fast algorithm to build word lists will increase performance and consistency of WEAT.

As part of pre-processing, we sort the sentences in our dataset by gender based on pronoun use and presence of male or female first names. We then create a lexical histogram from each of the two gendered sentence groups, which we use as input to Monroe et.al.'s (2009) weighted log-odds ratio algorithm (WLOR) (Liberman, 2014). Most first-order comparisons between two contrasting datasets estimate word usage rates, without considering rare words. WLOR accounts for this common mistake, with a null hypothesis that both lexical histograms being compared are making random selections from the same vocabulary. In our implementation of the algorithm, we use three lexical histograms as input: source X (word-list derived from male-subject sentences), source Y (word list derived from female-subject sentences), and some relevant background source Z (word list derived from entire case law database). The output is a word list and each word's score, which is the 'weighted log odds ratio', where positive values indicate that the word is favored by male sentences, and negative that the word is favored by female sentences. Words with a score near zero are about equally important to male and female sentences.

## 4.4 WLOR Output, Thematic Grouping

WLOR's output is a word list, but the words are not grouped by category. In order to use WLOR output as input to WEAT, we take two steps. First, we isolate the 500 most gendered words in CAP, meaning the 250 highest scoring words (most skewed towards men) and the 250 lowest scoring words (most skewed towards women). Second, we manually group those 500 words by category. After grouping, we have twelve categories of word lists representing biases in CAP. This process of categorizing the most skewed words resulted in the employment and family categories containing the largest list of words.

## 4.5 WLOR Output, K-Means Grouping

Our second approach categorizes the WLOR output automatically, using the clustering algorithm K-Means. K-Means clustering is a method of vector quantization that aims to partition 'observations' into k clusters. In this case, the 'observations' are vectors representing words. Each observation, or vector, belongs to the cluster with the nearest mean. Since word embedding algorithms represent words as vectors whose positions relative to each other represent the words' semantic and physical relationships in text, k-means clustering is a relatively effective method of topically clustering corpora. We therefore train word embedding algorithm Word2Vec on CAP and run the SciKitLearn implementation of K-Means on the resulting embedding. As post-processing, we filter the resulting clusters to only contain the 500 most male- and female scoring words from the WLOR output. We filter in this way because K-Means outputs all categories in a text, not just categories that are potentially biased or gender related. The overall K-Means cluster results might or might not have a bias, but the words within them are not necessarily gendered. This could lead to the same inconsistency as previous work.

### 4.6 WEAT

We train a word2Vec model on CAP in order to test both methods of word list generation as inputs for WEAT. To assign a magnitude to a given bias, we average the cosine similarity between the vectors of each word within the bias's word list to male and female terms. The cosine similarity represents the strength of the association between two terms.

### 4.7 Robustness Measures

We compare the two grouping methods against popular bias word lists used in previous work using Zhang et. al's consistency tests (2019). Their research shows that the measure of bias between different pairs of gendered words, such as (she, he) versus (him, her), or even different capitalizations of the same gender pair, and a single word often have different vector directions. This proves that arbitrarily-built clusters are not consistent inputs to WEAT. They further show that words within the same bias word list, such as 'job' versus 'salary,' and the same gender pair, such as she/he can produce different bias magnitudes and even different vector directions. For example, 'job' might skew towards men, while salary skews towards women. The problem here is obvious. Zhang et al. term this inconsistency between different gender pairs and word lists 'base pair stability.' We test our bias category word lists (the output of WLOR categorized thematically, and the K-Means clustered output) for base pair stability, following Zhang et al. We then compare our outputs' stability against bias category word lists popularly used in earlier research. We find that both our categorization techniques pass the base pair stability test, but bias category word lists used in other research do not.

Furthermore, previous work often discusses 'positive bias results', indicating that there is some amount of association between a gender and a bias categories. 'Positive bias results' are defined as any association between a given word list and a gender term, such as a pronoun. However, to our knowledge there is no discussion in previous work of the significance of bias magnitude. For example, 'employment' might have a bias against women with a magnitude of 0.2. But is 0.2 significant? How much more biased is a bias category with a magnitude of 0.4? The magnitude is meaningless without the understanding of significance. As explained above, WEAT measures bias by comparing the cosine similarity between two groups of vectors; but any threshold of similarity is deemed as 'bias.' To control for that potential pitfall, we determine the significance of WEAT output's magnitude by estimating the mean and standard deviation of gender bias in the embedding space: we analyze the gender bias of the 20,000 most frequent words in the embedding vocabulary, which is approximately normally distributed, and determine that a "significant" change in magnitude is a change of at least one standard deviation above the mean.

### 4.8 Comparison Over Time

Our research shows two new methods of identifying word lists representing bias in text. When used in WEAT, these word lists uncover significant gender bias in case law. Yet CAP spans three centuries; it is not surprising that gender biases exist, considering historical gender gaps. For example, women were not granted the right to vote until 1920—nearly two centuries after our first case law document in CAP. In order to emphasize meaningful gender bias, we repeat our word list generation process for every five-year chunk of case law in the last 100 years, using data from the U.S labor census. We track the bias magnitude's progress over time. In order to compare against historical gender trends occurring at the same time period, we regress our results against the rise of women in the workforce in the last 100 years. We find that while there is significant gender bias generally in case law, the bias magnitude decreases at about the same rate as women's participation in the workforce increases.

## 5 Results

### 5.1 Overview of Previous Work

To set up our point of comparison for our own methods, we first run WEAT using word lists from two influential papers in NLP bias detection literature: Caliskan et al. (2017) and Garg et. al., (2018). We choose Caliskan's employment word set, which includes general employment terms.

*executive, management, professional, corporation, salary, office, business, career*

Figure 1: Caliskan employment terms.

As discussed, WEAT also requires gender pairs to perform the vector distance calculation. Rather than rely on male and female names, as Caliskan et al. did, we choose the broader pronoun category from Garg et. al. (table 1). As no explanation is given in either paper for the choice of words within the word lists, we have no reason to assume that comparing the two sets from different papers is problematic.

| Female Terms | Male Terms |
|---|---|
| She, daughter, hers, her, mother, woman, girl, herself, female, sister, daughters, mothers, women, girls, femen, sisters, aunt, niece, nieces | He, son, his, him, father, man, boy, himself, male, brother, sons, fathers, men, boys, males, brothers, uncle, uncles, nephew, nephews |

Table 1: Garg gender terms.

As an aside, we note that Garg et al.'s gendered terms (Table 1) are also family terms, which likely skews the vectors against employment terms for reasons other than just their gendered-ness.

Following the literature, we define gender bias in our embedding as:

$$bias = \frac{\sum_n \overline{\text{male word}}}{|N_{male}|} - \frac{\sum_n \overline{\text{female word}}}{|N_{female}|} \quad (1)$$

Caliskan's manually clustered word sets produce an embedding bias against Garg's female word list of -0.03 in CAP. Performing Zhang's base-pair stability test, we find that this method is inconsistent—many of the base pairs, when compared against the same given word in the set, produce vectors with different directions. Vector directions represent the direction of bias—either towards men or women. Further, the slant of Garg's gender term list against Caliskan's employment terms do correspond to a known bias against women, but there is no discussion of "significance" of the magnitude of bias, making results difficult to analyze. We determine magnitude change significance ourselves by estimating the mean and standard deviation of gender slant in the embedding space (Table 2).

Based on the standard deviation of approximately 0.07 above an approximately -0.004 mean, we determine that although there is a slight preference for men over women in employment terms using the Caliskan-Garg employment bias word lists, it is less than one standard deviation below the mean and cannot be considered significant. Further, When we ran the data on a subsection of the word-set, the embedding bias direction shifted from biased against women, with a magnitude of -0.03, to a bias against men with a magnitude of 0.013. We determine that manual arbitrary clustering is not a robust test for gender bias.

| Mean | Standard Dev. |
|---|---|
| -0.0042 | 0.0738 |

Table 2: Mean and standard dev. in CAP.

### 5.2 WLOR Output, Thematic Grouping

We next run the WLOR algorithm on the full dataset. The word 'office' is the most male-skewed word in U.S case law in the last century, discounting male pronouns and legal terms. The word 'husband' is the most female-skewed word in U.S case law in the last century, discounting female pronouns. (As an aside, we note that there are no legal terms skewed towards women.)

| Female Words | Male Words |
|---|---|
| Husband, married, children, child, marriage, death, mother, daughter, divorce, unmarried | Office, pay, witness, company, goods, work, corporation, defendant, trial |

Table 3: Excerpt from the top twenty most important words for female and male subject sentences

We then isolate the 500 most gendered words in CAP, meaning the 250 highest scoring words (most skewed towards men) and the 250 lowest scoring words (most skewed towards women). We group the 500 terms thematically into word lists representing biases. The largest word lists represent employment and family. Although the words in Table 4 are sorted thematically, it is interesting to note that all employment terms came from the top 250 male-relating words. There were no employment terms in the top 250 female-skewing words. Similarly, all family terms came from the top 250 female skewed words, as there were no family terms in the top 250 male-skewed words.

| Female | Female | Female | Male | Male | Male |
|---|---|---|---|---|---|
| *prostitution, illicit, abortion, lewd, carnal, unchaste, seduced, bastard* | *Children, heirs, parents, parent, spouse, wife, husband, brother, sister, daughter* | *Incapable, sick, weak, feeble, mentally, physically, mental* | *Shot, fired, killed, drunk, shooting, fight* | *Price, amount, salary, penalty, cost, fine, prices* | *Engineer, foreman, employer, employment, contractor, master* |

Table 5: K-Means clustered (automatic grouping) WLOR sample.

After running the WLOR algorithm and creating the bias category word lists, we next determine our gender pronoun list for WEAT. We only include gender words in our male/female gender pair lists that are included in the top 500 most gendered words for men and women in the WLOR output. We do not include family terms in our base pair lists because of the potential bias against those words that are not gender-related. (For example, the word 'husband,' although facially male, is likely used in a family context. This is as opposed to he/she, which is used regardless of context.)

| Family Words | Employment Words |
|---|---|
| *Husband, baby, married, children, child, marriage, mother, father, divorce, unmarried, widow, wife, birth, divorced, family* | *Office, company, pay, goods, work, corporation, firm, business, engineer, employer, employment, employed, salary, client* |

Table 4: Excerpt of thematic grouping of highest-scoring WLOR results

We then input our word lists into WEAT in order to compute bias magnitude. Using the same gender slant definition and formula as in section 5.1, we calculate the bias of employment terms as -0.19 against women, and the bias of family terms as 0.22 against men. Based on the mean of -0.0042 and standard deviation of 0.0738 calculated for general gender slant in CAP, we find these results to be statistically significant.

Not only do the bias categories have statistically significant bias; each word within the bias categories has the same vector direction and are statistically significantly biased. This is different than previous research, whose word lists contained words with opposing vector directions. In order to determine this robustness, we perform Zhang's base-pair stability test by testing each word from within the same bias category separately against each set of gender pairs (such as she/he). We find that there is no directional change of vectors between different base-pairs and the same words. When testing each word separately against the she/he gender pair, both are independently biased towards women. Further, there is no significant change in bias magnitude (as defined by one standard deviation above the mean) between different words and base pairs. The results indicate that using first-order approaches, as we did with WLOR, is enough to identify basic categories of bias in a text, even if the output of the first-order method is manually grouped.

### 5.3 WLOR Output, K-Means Grouping

We next test to see if automatically clustering the WLOR output produces different results than the thematic grouping of WLOR output. The primary benefit of complete automatic clustering is that there is no "researcher bias", i.e., no assumptions of previous bias affect the clusters themselves. For example, in the manually-clustered WLOR output, we identified areas of bias by thematically grouping the output word list—but we still had an implicit awareness of the historical bias of men/work versus women/family. Automatic clustering frees the data entirely from researcher's potentially biased decision making. The drawback of this method is the heavy, slower Word2Vec training model.

We train a Word2Vec model on the entire dataset, and cluster the resulting embeddings using K-Means clustering with a preset of 300 clusters. We choose this algorithm for its speed and accuracy. In order to assess which clusters are gender related, we filter the resulting clusters to only include words in the WLOR output's 250 most male-skewed words and 250 most female-skewed words. This filtering controls the quality of the word lists: the word lists only contain words which already are known to be gendered in CAP. Upon visual inspection, most of the clusters seem relatively cohesive.

We use Zhang's base-pair stability test on all clusters with at least five words in the top 500 gendered words. There were seventeen clusters in this category. A sample of these can be seen in Table. 5. Interestingly, the resulting clusters primarily contain either male-skewed or female-skewed terms, but not both. All clusters that included primarily female-skewed terms were indeed found to be biased against women when used as inputs to WEAT. Similarly, all clusters with primarily male-skewed terms were found to be biased against men. Testing between each gender pair and each word in all seventeen clusters, we found that 97% of words within the same word list had the same vector direction. Sixteen out of the seventeen clusters produced had significant bias, meaning that the difference in gender slant scores was greater than, or less than, at least one standard deviation above or below the mean. We conclude that automatic clustering of first-order lexical histograms is a robust and consistent measurement of bias in text. We note that the automatic clustering also produced many categories of bias that we did not consider, such as associating demeaning sexual terms with women, and violence with men.

## 6 Comparison Over Time

We have shown that automating the formation of word lists to represent biases in WEAT leads to consistent and robust bias detection in text. Using two separate approaches, we created bias word lists to detect gender bias in case law and found that it exists. However, given the time span of our database, the presence of language difference between genders is not surprising.

In order to detect meaningful gender bias, i.e., bias that is stronger in text than real-world historical gender gaps, we track the change in bias magnitude over time. We regress the change in bias magnitude against women's participation in the workforce and find they progress at about the same rate.

### 6.1 Labor Slant

In order to compare the rate of change between gender bias in case law and women's participation in the workforce in the last 100 years, we first define the labor 'bias' for a given period in time. For precision, we label difference between men and women in the labor force as 'slant.' We define labor slant as the percentage of women in the workforce minus the percentage of men in the workforce. Formally:

$$slant = labor_{women} - labor_{men} \qquad (2)$$

The closer the labor slant is to zero, the more the workforce participation is equally divided between genders.

### 6.2 WLOR Results Over Time

We run the WLOR algorithm on each five-year slice of time in the last 100 years. The word lists generated from WLOR for female-subject sentences in the last century, discounting pronouns, include the word "husband" as the most important word consistently for every timeframe we analyzed between 1920 and 2017. The first five words for every five-year span in the last 100 years include the words "child/children", "mother", and "pregnant." The most consistently important words for male-subject sentences in the last century are "work", "guilty", and "sentence". Most words in the output generated for male-subject sentences mean "work", some kind of automobile, or are legal language.

This stark difference in language between two datasets separated by gender provides a clear picture of how the language used in our judicial system distinguishes between women and men: women are family oriented, and men are working (and driving, another form of mobility and therefore power) subjects of the law. The first time a family word appears in the male-subject list of important words was in 2005, with the word "wife." The first time an employment term appeared in the female-subject list of important words was an instance of the word "housework" in 1950. There are only three instances of employment terms for women between 1920 and 1990 out of 3,500 words analyzed in that time frame. It is also interesting to note the absence of legal language from the most heavily female words in the database. Although we do not explore this in our current research, we bring up the possibility that women are simply viewed as less meaningful subjects of law.

## 6.3 WEAT Results Over Time

We follow our two word-list building methods as inputs for WEAT for every five-year span of time in CAP in the last century. We use employment bias as our input wordlist for our WLOR thematic clustering approach, as the employment category has the largest word set. We find that for all years before 1980, words in our occupation-themed bias category are more associated with men than women, and after 1980, the trend hovers around 0, with a slight preference towards women. We use the cluster with primarily family terms as a 'family' bias as our input wordlist for K-Means, which is largest wordlist in our automatic approach. We find that there is a steady decrease in bias towards men in this category since 1920. We use the absolute values of these biases for clarity in our graph (Figure 2).

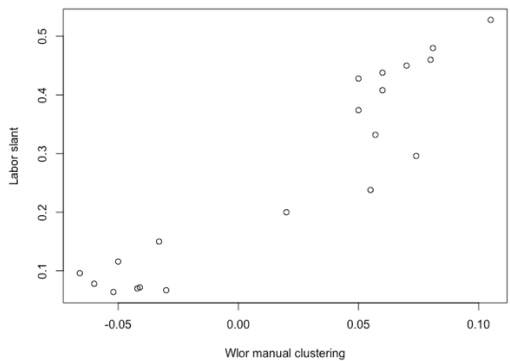

Figure 3: Labor slant and WLOR thematic clustering regression data.

## 6.4 Regression

We present results of each bias category word list regressed against the change in labor force participation in the last 100 years using data from the U.S census reports. We find that the change in bias magnitude over the last 100 years for both word lists are highly correlated with the increase of women in the workforce. Our results, with a P value of 1.172e-09 and an $R^2$ of 0.8781 for thematic grouping and a P value of 3.08e-09 and an $R^2$ of 0.8676, are consistent with the hypothesis that legal language's gender bias decreases as women's participation increases in the workforce.

## 7 Conclusion

In our research, we analyze 6.7 million documents of case law for gender bias. We find that existing bias detection methods in NLP are not sufficiently consistent to test for gender bias in CAP. We attribute this inconsistency to the lack of methodical building of word lists to represent bias. We therefore suggest two new approaches to building word lists for bias representation. We test our two approaches on CAP and find both methods to be robust and consistent, and to identify the presence of gender bias. We also show that, when the change in bias magnitude over time is regressed against workforce participation rate in the last 100 years, and find they are heavily correlated. It is worth noting that, although this research focuses specifically on gender bias, the same methodology might be applied to other groups—provided that those groups are identifiable in text.

As a future development in this research, we want to explore the results of our data that show that men are overwhelmingly associated with legal language, and women are not, even though women are not less likely to be defendants in certain types of law—such as Torts law (Chamallas, 2018). (In fact, Chamallas makes the point that Torts regulation can sometimes discriminate against women in other ways, and that Torts law should in fact have more female defendants than male.) Could it be that the law implicitly does not recognize women as independent legal entities in the same way it does men? We also would like to study possible intersections of identity in our judicial system. For example, we show that there is gender bias present in case law, but is there stronger bias against women of color than white women? Further, we wish to expand this research by involving judgments by experts of gender on developing a more holistic approach to bias clustering. Lastly, for future work we hope to analyze the impact these biases have on NLP systems' overall performance, and potential harms from these systems in other fields.

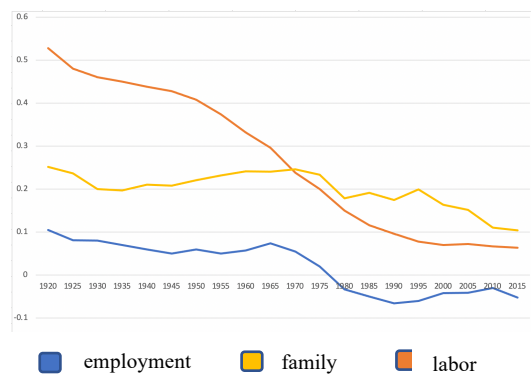

Figure 2: change in employment gender bias, family gender bias, and labor slant.

## Acknowledgements

This work is based on research conducted in the University of Pennsylvania's Cognitive Science and Computer Science departments under the guidance of Professor Mark Liberman. Thank you to Professor Liberman for code samples, editing, and advice. Thank you to Gedalia Gillis for statistical analysis, Professor Tom Baker (Penn Carey Law School) for help with legal analysis, Professor Talia Gillis (Columbia Law School) for the introduction to the CAP dataset, and Sharon Baker and anonymous reviewers for their helpful comments.